\documentclass{article}
\usepackage{ijcai19}

\usepackage{times}
\usepackage{soul}
\usepackage{url}
\usepackage[hidelinks]{hyperref}
\usepackage[utf8]{inputenc}
\usepackage[small]{caption}
\usepackage{graphicx}
\usepackage{amsmath}
\usepackage{booktabs}
\usepackage{algorithm}
\usepackage{url}
\usepackage{amssymb}
\usepackage{algpseudocode}
\usepackage{amsthm}
\usepackage{enumerate}
\newtheoremstyle{case}{}{}{}{}{}{:}{ }{}
\theoremstyle{case}

\theoremstyle{definition}

\newtheorem*{definition*}{Definition}

\newtheorem*{lemma*}{Lemma}
\usepackage{comment}
\usepackage{breakcites}

\title{Sampler for Composition Ratio by Markov Chain Monte Carlo}

\author{
Yachiko Obara \and Tetsuro Morimura \and Hiroki Yanagisawa
\affiliations
IBM Research - Tokyo\\
\emails
yachiko.obr@gmail.com,
\{tetsuro, yanagis\}@jp.ibm.com,
}

\begin{document}

\maketitle

\begin{abstract}
Invention involves combination, or more precisely, ratios of composition. According to Thomas Edison, {\em ``Genius is one percent inspiration and 99 percent perspiration''} is an example. In many situations, researchers and inventors already have a variety of data and manage to create something new by using it, but the key problem is how to select and combine knowledge. In this paper, we propose a new Markov chain Monte Carlo (MCMC) algorithm to generate composition ratios, nonnegative-integer-valued vectors with two properties: (i) the sum of the elements of each vector is constant, and (ii) only a small number of elements is nonzero. These constraints make it difficult for existing MCMC algorithms to sample composition ratios. The key points of our approach are (1) designing an appropriate target distribution by using a condition on the number of nonzero elements, and (2) changing values only between a certain pair of elements in each iteration. Through an experiment on creating a new cocktail, we show that the combination of the proposed method with supervised learning can solve a creative problem.
\end{abstract}

\section{Introduction}
The cocktail kir royal is a mixture of 90\% champagne and 10\% crème de cassis. Similarly, many elements of daily life can be regarded as mixtures, in which the amount of each component can be calculated as a ratio in relation to the whole: schedules, household expenses, investment portfolios, foods, drinks, cocktails, medicines, toiletries, cosmetics, fragrances, documents, if we do not consider the order. In this paper, a composition ratio means a set of such ratios in relation to the whole of a mixture's components. We denote a composition ratio as a nonnegative vector $\boldsymbol{x}$. The cocktail kir royal, for example, would be denoted as $\boldsymbol{x} = (90, 10, 0 ,\ldots, 0)$, where the first element of $\boldsymbol{x}$ is champagne, the second is crème de cassis, and the other elements represent other available ingredients. 
Composition ratios have two key characteristics. First, they are often sparse, containing a small number of components among a wide range of choices. Second, they have various desired properties. A fragrance, for example, is created by selecting dozens of ingredients from thousands of ingredients. 
The proportion of each selected ingredients is set under the condition that the total mass is $1000$ g, for example a fragrance composed of $700$ g of ``ingredient A'' and $300$ g of ``ingredient B''.
A fragrance can have desired properties related to aromatics (e.g., the type of smell), popularity (e.g., frequent patterns of ingredient combinations, or combinations that should be avoided), and appropriateness for certain use cases (e.g., combinations for perfumes, shampoos, or hand soaps). Perfumers who create new fragrances seek to develop various fragrances with desired properties. It is also possible that perfumers are willing to accept certain fragrances lacking some desired properties, because they can still draw inspiration from such fragrances. Thus, it is interesting to consider approaches for generating many fragrances in proportion to how well they satisfy desired conditions.

To solve such composition ratio problems, we propose a new Markov chain Monte Carlo (MCMC) algorithm and use MCMC samples themselves as solutions. MCMC is a strategy for generating samples $\boldsymbol{x}$ that follows samples drawn from a target distribution $P(\boldsymbol{x})$. 
The strategy works by using a Markov chain to spend more time in a state $\boldsymbol{x}$ that has a higher probability $P(\boldsymbol{x})$. MCMC techniques play fundamental roles in machine learning, physics, statistics, econometrics, and decision analysis \cite{andrieu2003introduction} and are often applied to solve integration problems in high-dimensional spaces for cases such as normalization, marginalization, and expectation. MCMC samples can also be used to obtain the maximum of an objective function, but this is inefficient as compared with other methods, including simulated annealing and gradient descent. Therefore, it is rare to use MCMC samples for the optimization problem, but we purposely use this approach because of the nature of our problem. 

The general approach to generating composition ratios having several desired properties can be seen as an optimization problem. The Markowitz standard model is a well-known model for optimizing investment portfolios \cite{markowitz1952portfolio}. Selecting sparse portfolios (i.e., portfolios with only a few active positions) is significant, because they allow accounting for transaction costs \cite{brodie2009sparse}. There are many improved approaches based on the Markowitz standard model to handle more complex conditions and a wider range of choices, such as a neural-network-based model \cite{fernandez2007portfolio} and a particle swarm optimization model \cite{cura2009particle}. Our proposed MCMC method does not overlap those methods but can be combined with them by appropriately designing the target distribution. In the same manner, we can combine our method with generative models, such as latent Dirichlet allocation (LDA) for handling documents as composition ratios \cite{blei2003latent}.

In our research, we focus on two aspects of the problem: creativity and resource allocation. For example, fragrance development emphasizes creativity, while investment portfolio selection emphasizes resource allocation. In this paper we address the former creative problem. Through an experiment on creating a new cocktail, we show that our method can solve such creative problems through combination with supervised learning.

The contributions of this paper are the following:
\begin{itemize}
	\item We propose an efficient MCMC algorithm to generate composition ratios with a small number of components chosen from a wide range of choices.
	\item We report empirical evidence that the combination of our method with supervised learning can solve a creative problem.
\end{itemize}

\section{Problem Formulation}\label{sec Problem Formulation}
In this section we formulate our problem. First, we define the composition ratio. In this paper, we consider discretized composition ratios. Specifically, a composition ratio is a nonnegative integer vector $\boldsymbol{x}$ having $N$ dimensions and the property that the sum of the elements of $\boldsymbol{x}$ is equal to an integer $M$. Thus, the set of composition ratios is
	%
	\begin{align}
		\label{eq:descritize}
    		{\mathcal{X} 
    			\triangleq  } \{ \boldsymbol{x} \in \mathbb{N}  _0^N 
			 	\mid  \Sigma_i x_i = M \}.
	\end{align}%
A vector $\boldsymbol{x} \in \mathcal{X}$ can be modeled as having $M$ balls in $N$ bins. The number of bins containing at least one ball is denoted as $n$. In the case of a cocktail, for example, $\boldsymbol{x}$ denotes the composition ratio of the cocktail, $N$ denotes the number of all available ingredients, $x_i$ denotes the amount of the $i$-th ingredient, and $n$ denotes the number of ingredients used in the cocktail. When $M$ equals 100, each $x_i$ can be regarded as the percentage of the $i$-th ingredient in relation to the whole. 

We consider the problem of generating $T$ random samples \{$\boldsymbol{x}^{(1)},$ $\boldsymbol{x}^{(2)},\ldots,$ $\boldsymbol{x}^{(t)},\ldots,$ $\boldsymbol{x}^{(T)} \}$ 
from a given probability distribution $P(\boldsymbol{x} | \mathcal{Y})$, where $\boldsymbol{x}^{(t)} \in \mathcal{X}$ and $\mathcal{Y} \triangleq \{y_1, y_2, ..., y_K\}$ is a set of conditions, which is given by user according to the application.  
We define 
	\begin{align}
		P(\boldsymbol{x} | \mathcal{Y}) 
		\triangleq \frac{\exp \{ -\Sigma_{k = 1}^{K} \mathop{E} (\boldsymbol{x} | y_k) \}}
		{\Sigma_{\boldsymbol{x}} \exp 
		\{ -\Sigma_{k=1}^{K} \mathop{E} (\boldsymbol{x} | y_k) \}}, 
	\label{equation:energy function}
	\end{align}%
where $\mathop{E}$ is a scalar-valued function called {\em the energy function}. 
Equation (\ref{equation:energy function}) is a common way to define a valid probability (because the value is always positive and the sum of all elements is one)  \cite{lecun2006tutorial}. 
We can define $P(\boldsymbol{x} | \mathcal{Y})$ indirectly by defining the energy function $E(\boldsymbol{x} | y_k)$. 
We also write the probability distribution $P(\boldsymbol{x} | y_k)$ $\propto$ 
$\exp \{- \mathop{E} (\boldsymbol{x} | y_k)\}$.  
Note that $P(\boldsymbol{x} | \mathcal{Y}) \propto  \prod_{k=1}^{K} P(\boldsymbol{x} | y_k)$. 

In this paper, we handle two types of conditions $y_1$ and $y_k (k>1)$. The condition $y_1$ describes the sparsity condition. The other $y_k (k>1)$ describe conditions of the other desired properties. For readability, we denote $y_1$ as $y_{\text{sparse}}$, and $y_k (k>1)$ as $y_{\text{property}}$ from here.
First, we define $y_{\text{property}}$ and its energy function. Let $y_{\text{property}}$ be a nonnegative-valued function of $\boldsymbol{x}$, which outputs the goodness of fit of $\boldsymbol{x}$ with respect to the targeted property.  The corresponding energy function is naturally defined as
	\begin{align}
	E (\boldsymbol{x} | y_{\text{property}} ) \triangleq - c_{\text{property}} \log y_{\text{property}} (\boldsymbol{x}), 
	\label{equation:y property}
	\end{align}%
where $c_{\text{property}} \in \mathbb{R}_{\geq 0}$ is a hyper-parameter controlling the priority of the condition $y_{\text{property}}$.
In the case of a cocktail, for example, a condition $y_{\text{property}}$ might demand that the taste of $\boldsymbol{x}$ should be ``Fresh''. As the taste of $\boldsymbol{x}$ is closer to ``Fresh'', $y_{\text{property}}(\boldsymbol{x})$ outputs a higher value.

As a second type of condition, we consider the conditions on the sparsity of the desired samples.  
We define the sparsity condition $y_{\text{sparse}}$ as the parameter of the categorical distribution, $y_{\text{sparse}} \in \mathbb{R}^N_{\geq0}$. The sparsity condition $y_{\text{sparse}}$ requires that $P(x|y_{\text{sparse}})$ satisfy
	\begin{align}
    		y_{\text{sparse}}(n) = \sum_{\boldsymbol{x}} P(\boldsymbol{x}|y_{\text{sparse}}) \mathbb{I}(\| \boldsymbol{x}\|_0=n),
		\label{equation yc}
	\end{align}%
where $\mathbb{I}$ is an indicator function, 
and $\| \boldsymbol{x}\|_0$ denotes $\ell_0$-norm of $\boldsymbol{x}$ (i.e., the number of nonzero elements of $\boldsymbol{x}$).
We derive the energy function $\mathop{E} (\boldsymbol{x} | y_{\text{sparse}})$ by solving equation~(\ref{equation yc}) in Section \ref{subsec How to handle the sparsity condition}. 

Before proceeding, we show examples of a sparsity condition $y_{\text{sparse}}$. In the case of creating cocktails, for example, $y_{\text{sparse}}$ indicates how many ingredients are likely to be used in a cocktail. There are several possible candidates for $y_{\text{sparse}}$. It can be a unimodal distribution if there is a rough desirable number of ingredients used in a cocktail. In another case, it can be a bimodal distribution when we aim to simultaneously generate simple cocktails (smaller number of ingredients) and complex cocktails (higher number of ingredients). Also, when a smaller number of ingredients is acceptable, we can use a distribution in which the probability exponentially decays as the number of ingredients increases. 
\section{MCMC Algorithms}\label{subsec MCMC algorithm}
An MCMC sampler is a standard algorithm that uses a Markov chain to generate samples from a target probability distribution $P(\boldsymbol{x})$. In Section \ref{subsec Markov Chain and MCMC} we start by reviewing Markov chains and MCMC methods. Section \ref{subsec Metropolis-Hastings algorithm} explains the Metropolis-Hastings algorithm \cite{Hastings1970} \cite{metropolis1953equation}, a notable MCMC algorithm. 
\subsection{Markov Chain and MCMC}\label{subsec Markov Chain and MCMC}
We consider a Markov chain, in which the future state $\boldsymbol{x}'$ depends only on the current state $\boldsymbol{x}$, which is a real-valued vector having $N$ dimensions. Let $\pi (\boldsymbol{x}, \boldsymbol{x}')$ be the transition probability of the Markov chain from $\boldsymbol{x}$ to $\boldsymbol{x}'$. It is known that, for any initial state, a Markov chain converges to an invariant distribution as long as the transition probability $\pi (\boldsymbol{x}, \boldsymbol{x}')$ satisifes the detailed balance condition, as follows. 
   	\theoremstyle{definition}
	\begin{definition*}{}
		The {\em detailed balance condition} is a sufficient condition to ensure that a distribution $P(\boldsymbol{x})$ is the invariant distribution to which a Markov chain converges, when the chain is ergodic. The condition is met when any $\boldsymbol{x}$ and $\boldsymbol{x}'$ satisfy the following condition: 
   		\begin{eqnarray}
   			P(\boldsymbol{x}) \pi (\boldsymbol{x}, 
				\boldsymbol{x}') = 
   				P(\boldsymbol{x}') \pi (\boldsymbol{x}', \boldsymbol{x}), \nonumber
			\label{equation:Detailed balance}
		\end{eqnarray}
	\end{definition*} 
Let $P(\boldsymbol{x})$ be a target distribution. Then, an MCMC sampler is a Markov chain that has 
an invariant distribution equal to the target distribution $P(\boldsymbol{x})$.
It is common in MCMC algorithms to design the transition probability $\pi(\boldsymbol{x},\boldsymbol{x}')$ to ensure that the detailed balance condition is satisfied. 
Note that each sample generated from an MCMC sampler can be seen as a visited state at each time step of the Markov chain, so we use the words ``sample" and ``state" interchangeably throughout this paper. 
To solve the problem formulated in the previous section: generating $T$ random samples \{$\boldsymbol{x}^{(1)},$ $\boldsymbol{x}^{(2)},\ldots,$ $\boldsymbol{x}^{(t)},\ldots,$ $\boldsymbol{x}^{(T)} \}$ 
from a given probability distribution $P(\boldsymbol{x} | \mathcal{Y})$, we use an MCMC algorithm. Instead of sampling each $\boldsymbol{x}$ from scratch, we sample $\boldsymbol{x}'$ according to the previous sample $\boldsymbol{x}$. Note that an MCMC algorithm assumes that the initial sample $\boldsymbol{x}^{(0)}$ is given. We sample $\boldsymbol{x}^{(1)}$ from $\boldsymbol{x}^{(0)}$ based on the transition probability $\pi (\boldsymbol{x}^{(0)}, \boldsymbol{x}^{(1)})$ and we repeat this process until we get $T$ random samples. 

\subsection{Metropolis-Hastings Algorithm}\label{subsec Metropolis-Hastings algorithm}
The Metropolis-Hastings algorithm is one of the most popular MCMC algorithms, and Algorithm \ref{alg:MH} lists its pseudocode. We denote the continuous uniform distribution of which minimum value is zero and maximum value is one as $\mathcal{U} (0,1)$. 
The Metropolis-Hastings algorithm requires the proposal distribution $Q(\boldsymbol{x}, \boldsymbol{x}')$ that the user should design in advance. 
In this algorithm a candidate $\boldsymbol{x}'$ for the next sample is drawn from a proposal distribution $Q(\boldsymbol{x}, \boldsymbol{x}')$, which is a probability distribution given the current sample $\boldsymbol{x}$. Depending on the acceptance rate $a(\boldsymbol{x}, \boldsymbol{x}')$ below, the candidate is accepted and used as the next sample. 
%
	\begin{align}\label{equation:acceptance rate}
 		a(\boldsymbol{x}, \boldsymbol{x}')  = 
		\min \left\{1, \frac{P(\boldsymbol{x}')
				Q(\boldsymbol{x}',\boldsymbol{x})}
				{P(\boldsymbol{x})
					Q(\boldsymbol{x},\boldsymbol{x}')} 
			\right\}. 
	\end{align}
When the candidate is rejected, the current sample is used as the next sample. In other words, the Markov chain moves from the current state $\boldsymbol{x}$ to the next state $\boldsymbol{x}'$ with acceptance rate $a(\boldsymbol{x}, \boldsymbol{x}')$, and otherwise, it remains at $\boldsymbol{x}$. The transition probability is calculated as $\pi (\boldsymbol{x}, \boldsymbol{x}')$ $= Q(\boldsymbol{x}, \boldsymbol{x}' )a(\boldsymbol{x}, \boldsymbol{x}')$. The detailed balance condition is known to be satisfied when the acceptance rate is used \cite{chib1995understanding}.
One drawback of the Metropolis-Hastings algorithm is that, if the acceptance rate $a(\boldsymbol{x},\boldsymbol{x}')$ is much smaller than one, it rejects many samples, the mixing time to invariant distribution is very high, and thus the computation time will be unacceptable.
It is important to design an appropriate proposal distribution so that the acceptance rate is close to one.
   	\begin{algorithm}[t]
		\caption{Metropolis-Hastings Algorithm}
		\label{alg:MH}
		\textbf{Input}:  $N$, $\boldsymbol{x}^{(0)}$, $T$, $Q(\boldsymbol{x}, \boldsymbol{x}')$ ,$P(\boldsymbol{x})$\\
		\textbf{Output}: $\boldsymbol{x}^{(1)}, \ldots, \boldsymbol{x}^{(T)}$
		\begin{algorithmic}[1] 
			\For{$t \gets$ 1 to $T$}
				\State $\boldsymbol{x}' \sim $ $Q(\boldsymbol{x}^{(t-1)}, \boldsymbol{x}')$
				\State $ a(\boldsymbol{x}^{(t-1)}, \boldsymbol{x}')  \gets 
					\min \{1, 
					\frac{P(\boldsymbol{x}')
					Q(\boldsymbol{x}',\boldsymbol{x}^{(t-1)})}
						{P(\boldsymbol{x}^{(t-1)})
						Q(\boldsymbol{x}^{(t-1)},
						\boldsymbol{x}')}\} $ \\
					\Comment{Acceptance rate}
				\State $u \sim \mathcal{U} (0,1)$
				\If {$u < a(\boldsymbol{x}^{(t-1)}, \boldsymbol{x}')$}
					\State $\boldsymbol{x}^{(t)} \gets
					\boldsymbol{x}'$
					 \Comment{Accept proposal}
				\Else
					\State $\boldsymbol{x}^{(t)} \gets 
					\boldsymbol{x}^{(t-1)}$
					\Comment{Reject proposal}
				\EndIf
			\EndFor
			\State \textbf{return} $\boldsymbol{x}^{(1)}, \ldots, \boldsymbol{x}^{(T)}$
		\end{algorithmic}
	\end{algorithm}
\section{MCMC Algorithm for Composition Ratios}
\label{sec MCMC Algorithm for Composition Ratio}
In this section we propose our MCMC algorithm for composition ratios. We start in Section \ref{subsec How to handle the sparsity condition} by describing how to handle the sparsity condition $y_{\text{sparse}}$. Section \ref{Naive MCMC Algorithm} shows that a naive MCMC algorithm based on the Metropolis-Hastings algorithm has a small acceptance rate, and we thus propose a new algorithm to improve the acceptance rate in Section \ref{Accelerated MCMC Algorithm for Composition Ratios}. In this section, for simplicity, we sometimes omit $\mathcal{Y}$ from the target distribution $P(\boldsymbol{x} | \mathcal{Y})$.  
Specifically, $P(\boldsymbol{x} | \mathcal{Y})$ is sometimes denoted as $P(\boldsymbol{x})$.
\subsection{Energy Function of Sparsity Condition}
\label{subsec How to handle the sparsity condition}
To handle the sparsity condition $y_{\text{sparse}}$, we derive the energy function $E (\boldsymbol{x} | y_{\text{sparse}})$ by solving equation~(\ref{equation yc}). As the sparsity condition $y_{\text{sparse}}$ relates only to the number of nonzero elements $n$, an $\boldsymbol{x}$ having the same number of nonzero elements $n$ has the same probability $P(\boldsymbol{x} | y_{\text{sparse}})$, so equation~(\ref{equation yc}) can be rewritten as the following: 
	\begin{align}
		P(\boldsymbol{x} | y_{\text{sparse}})
		= \frac{y_{\text{sparse}} (n)}{\Sigma _{\boldsymbol{x}' \in \mathcal{X}} 
		\mathbb{I}(\| \boldsymbol{x}'\|_0=n)}, 
		\label{equation:rewritten yc}
	\end{align}%
where the denominator is the number of $\boldsymbol{x} \in \mathcal{X}$ that satisfy $\| \boldsymbol{x}\|_0 = n$. From equation~(\ref{equation:rewritten yc}) and the definition of $P(\boldsymbol{x} |y_{\text{sparse}} )$ $\propto$ $\exp \{- \mathop{c}_{\text{sparse}} \mathop{E} (\boldsymbol{x} | y_{\text{sparse}})\}$ (cf. equation (\ref{equation:energy function})), the energy function $E(\boldsymbol{x} | y_{\text{sparse}})$ can be derived as 
	%
	\begin{eqnarray}
    	E (\boldsymbol{x} | y_{\text{sparse}})
    	&=& - c_{\text{sparse}} \log P(\boldsymbol{x} | y_{\text{sparse}}) \nonumber \\
	&=& - c_{\text{sparse}} \log \frac{y_{\text{sparse}} (n)}{\Sigma _{\boldsymbol{x}' \in \mathcal{X}} 
		\mathbb{I}(\| \boldsymbol{x}'\|_0=n)}, 
		\label{equation:sparse energy}
	\end{eqnarray}%
where $c_{\text{sparse}} \in \mathbb{R}_{\geq 0}$ is the hyperparameter controlling the priority of the condition $y_{\text{sparse}}$. In this paper we set $c_{\text{sparse}} =1$. The denominator of equation~(\ref{equation:sparse energy}) can be computed as
   %
	\begin{align}
    		\Sigma _{\boldsymbol{x}' \in \mathcal{X}} \mathbb{I}
			(\| \boldsymbol{x}'\|_0=n)
			    = {N \choose n}{M-1 \choose M-n},
	\label{equation combinations}
	\end{align}%
The first term on the right-hand side of equation~(\ref{equation combinations}) corresponds to the number of combinations for choosing $n$ bins from $N$ bins. The second term corresponds to the number of combinations that put a ball in each of the chosen $n$ bins and choose bins for allocating the remaining $M-n$ balls from $n$ bins. There are more combinations in the case in which the balls are distributed in many bins, so with the balls allocated randomly, $n$ tends to become larger, as listed in Table~\ref{table:combinations}. As described in the next section, our algorithm avoids directly calculating the number of combinations in equation~(\ref{equation combinations}).
   %
	\begin{table}[t]
		\centering
		\begin{tabular}{lrr}
			\toprule
			$n$  & The number of combinations  \\
			\midrule
			1 & 50 \\
			2 & $4\,900$  \\
			3 & $117\,600$  \\
			4 & $921\,200$ \\
			5 &$2\,118\,760$ \\
			\bottomrule
		\end{tabular}
		\caption{The number of combinations when we have $M=5$ balls and $N=50$ bins.}\label{table:combinations}
	\end{table}
\subsection{Naive MCMC Algorithm}\label{Naive MCMC Algorithm}
As a naive MCMC algorithm, we first describe the case of using the Metropolis-Hastings algorithm to choose a candidate for the next sample $\boldsymbol{x}'$ from a set of composition ratios $\mathcal{X}$ uniformly at random. This is achieved by setting the proposal distribution to the following: 
	\begin{eqnarray}
		Q(\boldsymbol{x}, \boldsymbol{x}') 
			= \begin{cases}
				\frac{1}{|\mathcal{X}|}
				& \text{if } 
				\boldsymbol{x}' \in \mathcal{X}\\
				0
				& \text{otherwise,}
		 \end{cases} 
		 \label{equation:naive}
	\end{eqnarray}%
where $|\mathcal{X}|$ is the number of elements of $\mathcal{X}$. By equation~(\ref{equation:acceptance rate}), the acceptance rate is set to $a(\boldsymbol{x}, \boldsymbol{x}') = \min \{1, P(\boldsymbol{x}')/P(\boldsymbol{x})\}$. 
For simplicity, we assume that $P(\boldsymbol{x}|\mathcal{Y})=P(\boldsymbol{x}|y_{\text{sparse}})$. Here, the acceptance rate tends to be small, because non-sparse samples having low probability $P(\boldsymbol{x})$ are drawn more often than sparse samples having high probability $P(\boldsymbol{x})$ (see Table~\ref{table:combinations}). Designing the proposal distribution to achieve a high acceptance rate under the sparsity condition is a nontrivial problem. Hence, the next subsection describes how to improve the acceptance rate.
\subsection{Accelerated MCMC Algorithm for Composition Ratios}\label{Accelerated MCMC Algorithm for Composition Ratios}
To increase acceptance rate, we propose a new MCMC algorithm.  
Algorithm \ref{alg:accelerated} lists the pseudocode for this algorithm. We denote the discrete uniform distribution on integers of which minimum value is one and maximum value is $N$ as $\mathcal{U} \{1,N\}$. 
Instead of sampling each $\boldsymbol{x}'$ from the proposal distribution $Q(\boldsymbol{x},\boldsymbol{x}')$ given in equation~(\ref{equation:naive}), we sample $\boldsymbol{x}'$ according to the previous sample $\boldsymbol{x}$. 
In each iteration we change values of only two elements, $x_i$ and $x_j$, of the previous sample $\boldsymbol{x}$, meeting the condition of equation~(\ref{eq:descritize}). 
We introduce a constraint on the probability for choosing the pair of elements to be changed in each iteration, namely, that at least one of the elements is nonzero. 
This constraint avoids the case in which both $x_i$ and $x_j$ are zeros, causing $\boldsymbol{x}'$ to remain $\boldsymbol{x}$. Such cases often occur because $\boldsymbol{x}$ is sparse, and this is why we introduce the above constraint. 
We denote the probability for choosing the pair of elements to be changed as $\alpha_{ij}(\boldsymbol{x})$. 
The constraint is described that we set $\alpha_{ij}(\boldsymbol{x})=0$ for most pairs of $i$ and $j$. 
Furthermore, instead of changing the values of $x_i$ and $x_j$ uniformly at random, we change them based on the probability which is proportional to conditions of the target distribution. 
Hence, we define the proposal distribution as 
	\begin{align}
		Q(\boldsymbol{x}, \boldsymbol{x}') 
			= \begin{cases}
				\alpha_{ij}(\boldsymbol{x}) P(x_i', x_j' | \boldsymbol{x}_{/ij})
				& \text{if } 
				\boldsymbol{x}_{/ij} =  \boldsymbol{x}'_{/ij}\\
				0
				& \text{otherwise,}
		 \end{cases} 
	\end{align}%
where $\boldsymbol{x}_{/ij}$ denotes $x_1, \ldots ,x_{N}$ but with $x_i$ and $x_j$ omitted. 
The idea of using conditions of the target distribution as the proposal distribution is known as Gibbs sampling \cite{geman1984stochastic}. 
In Gibbs sampling $\alpha_{ij}(\boldsymbol{x})$ is a constant (not depends on $\boldsymbol{x}$), and so the acceptance rate is one. 
We introduce the constraint that $\alpha_{ij}(\boldsymbol{x})$ depends on $\boldsymbol{x}$, and we can offset the effect of the constraint by adjusting the MCMC acceptance rate $a(\boldsymbol{x}, \boldsymbol{x}')$ in equation~(\ref{equation:acceptance rate}). 
Thus, we use the following adjusted acceptance rate: 
	\begin{align}\label{equation:acceptance rate2}
 		a(\boldsymbol{x}, \boldsymbol{x}')  = 
		\min \left\{1, \frac{\alpha_{ij}(\boldsymbol{x}')}{\alpha_{ij}(\boldsymbol{x})} 
			\right\}.
	\end{align}
This is derived by using $P(x_i, x_j | \boldsymbol{x}_{/ij})P(\boldsymbol{x}_{/ij}) = P(\boldsymbol{x})$, 
where $P(\boldsymbol{x}_{/ij})$ is the marginal distribution of $P(\boldsymbol{x})$ marginalized with respect to $x_i$ and $x_j$. 
The proposed algorithm satisfies the detailed balance condition, which can be shown in the same manner as for the Metropolis-Hastings algorithm. 

Here, we consider equation~(\ref{equation:acceptance rate2}) in detail to show that the acceptance rate can be small. 
In Algorithm \ref{alg:accelerated} we set $\alpha_{ij}(\boldsymbol{x})$ as the following: 
	\begin{eqnarray}
		\alpha_{ij}(\boldsymbol{x})
			= \begin{cases}
				\frac{2}{n(N-1)}
				& \text{if } 
				x_i > 0, x_j > 0, i>j, \\
				\frac{1}{n(N-1)}
				& \text{if } 
				x_i > 0, x_j = 0\\
				0
				&  \text{otherwise.} 
		 \end{cases} \nonumber
		 \label{equation alpha}
	\end{eqnarray}%
In each iteration, $x_i+x_j$ balls are reallocated into $x_i'$ and $x_j'$. Therefore, the number of nonzero elements, $n'$, of sample $\boldsymbol{x}'$ satisfies $n-1 \leq n' \leq n+1$, where $n$ is the number of nonzero elements of the previous sample $\boldsymbol{x}$: 
\begin{eqnarray}\label{equation:acceptance rate3}
 		\frac{\alpha_{ij}(\boldsymbol{x}')}{\alpha_{ij}(\boldsymbol{x})} 
		= \begin{cases}
  			\frac{2}{1+1/n}, & \text{if $n' = n+1$},\\
  	 		\frac{1}{2(1-1/n)}, & \text{if $n' = n-1$},\\
  			1,            & \text{if $n' = n$}.
 		 \end{cases} 
	\end{eqnarray}
Therefore, when the number of nonzero elements decreases, the acceptance rate can be small. 
We also confirm that the acceptance rate is always greater than 1/2 from equation~(\ref{equation:acceptance rate3}).

Next, we explain how the algorithm avoids calculating the number of combinations in equation~(\ref{equation combinations}). Instead of calculating $P(x'_i,x'_j | \boldsymbol{x}_{/ij}, y_{\text{sparse}}) \propto P(x'_i,x'_j | \boldsymbol{x}_{/ij}, \mathcal{Y})$ directly, we calculate the ratio of $P(x'_i,x'_j | \boldsymbol{x}_{/ij}, y_{\text{sparse}})$ and $P(x_i,x_j | \boldsymbol{x}_{/ij}, y_{\text{sparse}})$, i.e., the right-hand side of the following equation: 
	\begin{eqnarray}
     		P(x'_i,x'_j | \boldsymbol{x}_{/ij}, y_{\text{sparse}}) \propto
     		\frac{y_{\text{sparse}} (n') / {N \choose n'}{M-1 \choose M-n'}}
			{y_{\text{sparse}} (n) / {N 	\choose n}{M-1 \choose M-n}}. \nonumber
	\end{eqnarray}%
Notice that we have 
	\begin{eqnarray}
     		\frac{1 / {N \choose n'}{M-1 \choose M-n'}}
			{1/ {N \choose n}{M-1\choose M-n}} 
		 = \begin{cases}
  			\frac{n(n+1)}{(N-n)(M-n)},  & \text{if $n' = n+1$},\\
  	 		\frac{(N-n+1)(M-n+1)}{n(n-1)}, & \text{if $n' = n-1$},\\
  			1,            & \text{if $n' = n$}.
 		 \end{cases} \nonumber
	\end{eqnarray}%
Thus, we can avoid calculating the large number of combinations. 
	\begin{algorithm}[t]
		\caption{Accelerated MCMC for Sparse Composition Ratio}
		\label{alg:accelerated}
		\textbf{Input}: $N$, $\boldsymbol{x}^{(0)}$, $T$, $P(\boldsymbol{x})$\\
		\textbf{Output}: $\boldsymbol{x}^{(1)}, \ldots, \boldsymbol{x}^{(T)}$
		\begin{algorithmic}[1] 
			\For{$t \gets$ 1 to $T$}
				\State $i \sim $ $\mathcal{U} 
				\{k \mid x^{(t-1)}_k > 0,  k \leq N,
					 k \in \mathbb{N}  _0\}$
				\State $j \sim $ $\mathcal{U} \{1, N\}$ 
				\While{$i = j$}
					\State $j \sim $ $\mathcal{U} \{1, N\}$ 
				\EndWhile
				\State $x'_i,x'_j \sim  
					P(x'_i,x'_j | 
					\boldsymbol{x}^{(t-1)}_{/ij}$)
					\Comment{Proposal}
				\State $ a  \gets 
					\min \{1, 
					\frac{P(\boldsymbol{x}')
					Q(\boldsymbol{x}',\boldsymbol{x}^{(t-1)})}
						{P(\boldsymbol{x}^{(t-1)})
						Q(\boldsymbol{x}^{(t-1)},
						\boldsymbol{x}')}\} $ \\
					\Comment{Acceptance rate}
				\State $u \sim \mathcal{U} (0,1)$
				\If {$u < a$}
					\State $x_i^{(t)} \gets x'_i $ \Comment{Accept  proposal}
					\State $x_j^{(t)} \gets x'_j $
					\State $\boldsymbol{x}^{(t)}_{/ij}  
					\gets
					\boldsymbol{x}^{(t-1)}_{/ij}$
				\Else
					\State $\boldsymbol{x}^{(t)} \gets 
					\boldsymbol{x}^{(t-1)}$
					\Comment{Reject proposal}
				\EndIf
			\EndFor
			\State \textbf{return} $\boldsymbol{x}^{(1)}, \ldots, \boldsymbol{x}^{(T)}$
		\end{algorithmic}
	\end{algorithm}
\section{Experimental Results}
This section describes the results of two experiments. In the first experiment (Section \ref{subsec Control Sparsity}), we verified that samples drawn by the proposed MCMC algorithm satisfied a sparsity condition and converged to a target distribution. In the second experiment (Section \ref{subsec Creating a New Cocktail}), we attempted to use the algorithm in a creative work by creating a new cocktail. We discarded $10\,000$ iterations at the beginning of each MCMC run, which is a common practice to find a good starting point for MCMC methods.
\subsection{Control of Sparsity}\label{subsec Control Sparsity} 
We evaluated the proposed algorithm on the task of satisfying a sparse condition $y_{\text{sparse}}$. 
We conducted two experiments: one used small $N$ and small $M$  (a small ingredients set and a rough division), 
and another used large $N$ and large $M$ (a large ingredients set and a fine division). 
We also evaluated the naive MCMC algorithm and Gibbs sampling. 
Here Gibbs sampling is the same as Algorithm \ref{alg:accelerated} except $\alpha_{ij}(\boldsymbol{x})$ is constant (not depends on $\boldsymbol{x}$), 
so it satisfies the condition that the sum of the elements of each sample is constant. 

\subsubsection{Small Ingredients Set and Rough Division}\label{subsubsec Control Sparsity 1} 
We set $P(\boldsymbol{x}|\mathcal{Y}) = P(\boldsymbol{x}|y_{\text{sparse}})$, $N=10$, and $M=5$.
We conducted experiments on two target distributions: (a) a uniform distribution $y_{\text{sparse}}(n) = \mathcal{U} \{2, 5\}$ and (b) a unimodal distribution $y_{\text{sparse}}(n) \propto \exp(- 0.25(n-3)^2)$.
We ran $T = 50\,000$ iterations. 
For each target distribution, we sampled 10 times while changing the initial sample, which was generated randomly from $P(\boldsymbol{x}|\mathcal{Y})$. 

The both MCMC samples of the naive MCMC algorithm and Gibbs sampling did not match the target distributions while Algorithm \ref{alg:accelerated} converged. 
Table~\ref{table:Ex1mini} lists the accepted rates of proposed MCMC samples with the naive MCMC algorithm, Gibbs sampling, and Algorithm \ref{alg:accelerated}. 
We calculated how often the new sample $\boldsymbol{x}^{(t)}$ was different from the current sample $\boldsymbol{x}^{(t-1)}$ and listed them as updated rates in Table~\ref{table:Ex1mini updated}. 
Although the accepted rate of Gibbs sampling is one,  it is often the case in which both $x_i$ and $x_j$ are zero, causing a new sample $\boldsymbol{x}^{(t)}$ to remain the previous sample $\boldsymbol{x}^{(t-1)}$. Such cases often occur because $\boldsymbol{x}$ is sparse. 
Regarding the naive MCMC algorithm, it worked well on sampling various samples because both of the accepted rate and the updated rate are more than 70$\%$. 
However those samples did not match the target distributions. 

	\begin{table*}[t]
		\centering
		\begin{tabular}{lrrr}
			\toprule
			Target distribution  & Naive MCMC & Gibbs sampling &Algorithm \ref{alg:accelerated} \\ 
			  & accepted rates[$\%$] &accepted rates[$\%$] &accepted rates[$\%$]\\
			\midrule
			(a) Uniform  &70.40 $\pm$ 0.430 & 100.00 $\pm$ 0.000 & 95.58 $\pm$ 0.008 \\
			(b) Unimodal  &85.91 $\pm$ 0.437 & 100.00 $\pm$ 0.000 & 95.85 $\pm$ 0.001 \\
		\bottomrule
		\end{tabular}
	\caption{Accepted rates of proposed MCMC samples with the naive MCMC algorithm, Gibbs sampling, and Algorithm \ref{alg:accelerated}.}
	\label{table:Ex1mini}
	\end{table*}

	\begin{table*}[t]
		\centering
		\begin{tabular}{lrrr}
			\toprule
			Target distribution  & Naive MCMC & Gibbs sampling &Algorithm \ref{alg:accelerated}\\ 
			  & updated rates[$\%$] &updated rates[$\%$] &updated rates[$\%$]\\
			\midrule
			(a) Uniform  &70.32 $\pm$ 0.418 & 8.79 $\pm$ 0.001 & 53.53 $\pm$ 0.005\\
			(b) Unimodal  &85.90 $\pm$ 0.431 & 4.96 $\pm$ 0.002 & 57.37 $\pm$ 0.001\\
		\bottomrule
		\end{tabular}
	\caption{Updated rates of proposed MCMC samples with the naive MCMC algorithm, Gibbs sampling, and Algorithm \ref{alg:accelerated}.}
	\label{table:Ex1mini updated}
	\end{table*}

\subsubsection{Large Ingredients Set and Fine Division}\label{subsubsec Control Sparsity 2} 
We set $P(\boldsymbol{x}|\mathcal{Y}) = P(\boldsymbol{x}|y_{\text{sparse}})$, $N=2000$, and $M=100$. We conducted experiments on four target distributions: (a) a uniform distribution $y_{\text{sparse}}(n) = \mathcal{U} \{17, 24\}$, (b) a unimodal distribution $y_{\text{sparse}}(n) \propto \exp(- 0.25(n-20)^2)$, (c) a bimodal distribution $y_{\text{sparse}}(n) \propto \exp (-0.5(n-15)^2) + 2\exp (-0.5(n-20)^2)$ and (d) an exponential distribution $y_{\text{sparse}}(n) \propto 0.5\exp (-0.5n) $. We ran $T = 500\,000$ iterations. For each target distribution, we sampled 10 times while changing the initial sample, which was generated randomly from $P(\boldsymbol{x}|\mathcal{Y})$.

\begin{figure}[t]
		\includegraphics[trim={1.5cm 1cm 0 2cm},clip,,width=10.0cm]
		{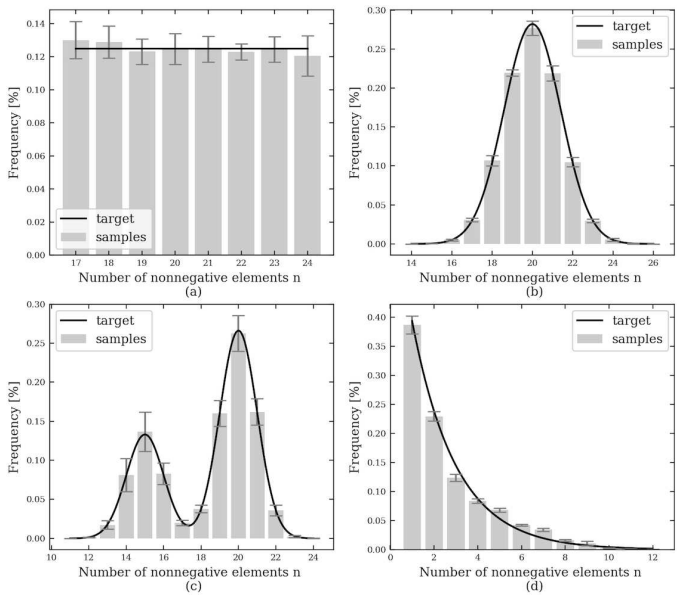}
		\caption{Experimental results with the target distribution defined as (a) a uniform distribution, (b) a unimodal distribution, (c) a bimodal distribution, and (d) an exponential distribution. In each graph, the curve shows the target distribution $y_{\text{sparse}}$, while the bars constitute a histogram of the mean number of nonnegative elements $n$ from MCMC samples of 10 sequences. Each error bar shows the standard deviation.}
		\label{fig:Ex1}
	\end{figure}

Figure \ref{fig:Ex1} shows histograms of the mean number of nonnegative elements $n$ from MCMC samples of 10 sequences. The histograms approximate the target distributions well. 
Table~\ref{table:Ex1} and Table~\ref{table:Ex1 updated} list the accepted rates and the updated rates of proposed MCMC samples with the naive MCMC algorithm and Algorithm \ref{alg:accelerated} respectively. 
Because $N$ was much larger than the expected $n$, the naive MCMC algorithm could not propose appropriate candidates, and so the acceptance rates were zero. We thus confirmed that the accepted rates with Algorithm \ref{alg:accelerated} were much larger than those with the naive MCMC algorithm. 
In the same manner as Gibbs sampling in the previous experiments, Gibbs sampling had small updated rates and did not match the target distributions. 
	\begin{table*}[t]
		\centering
		\begin{tabular}{lrrr}
			\toprule
			Target distribution  & Naive MCMC & Gibbs sampling &Algorithm \ref{alg:accelerated} \\ 
			  & accepted rates[$\%$] &accepted rates[$\%$] &accepted rates[$\%$]\\
			\midrule
			(a) Uniform  &0.00 $\pm$ 0.000 & 100.00 $\pm$ 0.000 & 99.60 $\pm$ 0.060 \\
			(b) Unimodal  &0.00 $\pm$ 0.000 & 100.00 $\pm$ 0.000 & 99.59 $\pm$ 0.012 \\
			(c) Bimodal &0.00 $\pm$ 0.000 & 100.00 $\pm$ 0.000 & 99.59 $\pm$ 0.010 \\
			(d) Exponential &0.00 $\pm$ 0.000 & 100.00 $\pm$ 0.000 & 99.97 $\pm$ 0.0002 \\
		\bottomrule
		\end{tabular}
	\caption{Accepted rates of proposed MCMC samples with the naive MCMC algorithm, Gibbs sampling, and Algorithm \ref{alg:accelerated}.}
	\label{table:Ex1}
	\end{table*}

	\begin{table*}[t]
		\centering
		\begin{tabular}{lrrr}
			\toprule
			Target distribution  & Naive MCMC & Gibbs sampling &Algorithm \ref{alg:accelerated}\\ 
			  & updated rates[$\%$] &updated rates[$\%$] &updated rates[$\%$]\\
			\midrule
			(a) Uniform  &0.00 $\pm$ 0.000 & 0.43 $\pm$ 0.016 &  50.31 $\pm$ 0.051\\
			(b) Unimodal  &0.00 $\pm$ 0.000 & 0.10 $\pm$ 0.070 & 50.32 $\pm$ 0.081\\
			(c) Bimodal &0.00 $\pm$ 0.000 & 0.14 $\pm$ 0.089  &  50.25 $\pm$ 0.055\\
			(d) Exponential &0.00 $\pm$ 0.000 & 0.03 $\pm$ 0.003 &  50.07 $\pm$ 0.010\\
		\bottomrule
		\end{tabular}
	\caption{Updated rates of proposed MCMC samples with the naive MCMC algorithm, Gibbs sampling, and Algorithm \ref{alg:accelerated}.}
	\label{table:Ex1 updated}
	\end{table*}
	
\begin{figure}[t]
		\includegraphics[trim={0 0 0 0},clip,,width=8.5cm]{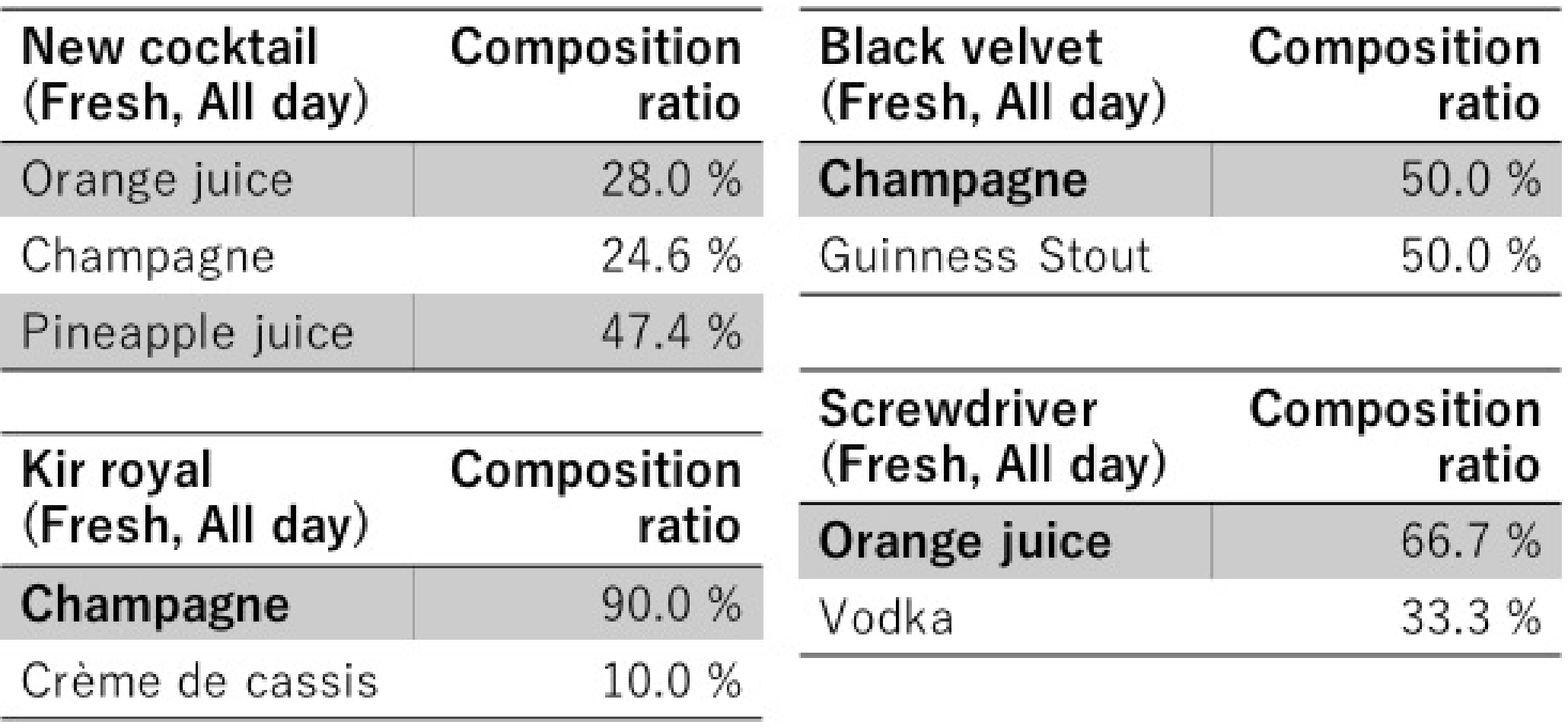}
		\caption{Recipes of cocktails including a new cocktail and the cocktails in the dataset 
		similar to the new cocktail. 
		The ingredients with bold show that they are common with the new cocktail.}
	\label{fig:cocktail_recipe}
	\end{figure}
%
\subsection{Creation of New Cocktail}\label{subsec Creating a New Cocktail}
We also evaluated the proposed algorithm on the task of creating new cocktails. We used a cocktail dataset \cite{Stevan2019} containing 69 cocktails, with seven taste labels $\{$``Bittersweet,'' ``Boozy,'' ``Fresh,'' ``Salty,'' ``Sour,'' ``Sweet,'' ``Unknown''$\}$, and four timing labels $\{$``After dinner,'' ``All day,'' ``Long drink,'' ``Pre-dinner''$\}$. We aimed to generate $100\,000$ cocktails (samples) labeled as ``All day'' and ``Fresh.'' The total number of ingredients, $N$, was $65$. We converted all ingredient unit amounts (e.g., centiliter, bar spoon, dash, splash) into permille values, giving $M = 1000$. In the dataset, the number of ingredients used in one cocktail ranged from two to five, and we set $y_{\text{sparse}}$ accordingly: $y_{\text{sparse}}(2) = 13/69, y_{\text{sparse}}(3) = 29/69, y_{\text{sparse}}(4) = 24/69, $ and $y_{\text{sparse}}(5) = 3/69$. 
We also add the conditions $y_{\text{taste}}$ and $y_{\text{timing}}$ 
demanding that the taste be ``Fresh'' and timing be ``All day", respectively. 
We trained two multi-class classification models using a random forest: one for the taste labels, and the other for the timing labels. 
We then used the corresponding outputs as $y_{\text{taste}}(\boldsymbol{x})$ and $y_{\text{timing}}(\boldsymbol{x})$. 
Note that the output of a random forest can be regarded as a probability \cite{bostrom2007estimating}. We set the hyperparameter $c$ to one and ran the proposed MCMC algorithm for $T = 110\,000$ iterations. The initial sample was generated randomly under the condition of $y_{\text{sparse}}$.

\begin{figure}[t]
		\includegraphics[trim={0 0 0 0},clip,,width=9.0cm]{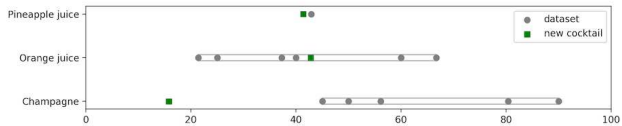}
		\caption{Composition ratios of the new cocktail and similar cocktails in the dataset.}
	\label{fig:new_cocktail}
	\end{figure}

Figure \ref{fig:cocktail_recipe} shows recipes (i.e., composition ratios) for a new cocktail and the most similar cocktails in the dataset. To calculate similarities, we focused on combinations of ingredients and used the Szymkiewicz-Simpson coefficient \cite{simpson1960notes}. All the most similar cocktails were labeled with ``All day'' and ``Fresh,'' but they had only one ingredient in common with the new cocktail. This result suggests that we should have the classification models focus on co-occurrence to get improved outcomes. The combination of champagne and orange juice used in the new cocktail did not appear in the dataset but is actually a popular cocktail called a mimosa. Our algorithm thus reinvented the combination of ingredients in a popular cocktail. Figure \ref{fig:new_cocktail} shows the composition ratios of the new cocktail and the similar cocktails in the dataset. The composition ratio of champagne is smaller in the new cocktail than in the others. We attributed this to the constraint imposed by the composition ratios with $M=1000$, but this problem requires more investigation.

\begin{figure}[t]
		\includegraphics[trim={1.5cm 0 0 1cm},clip,,width=10.0cm]{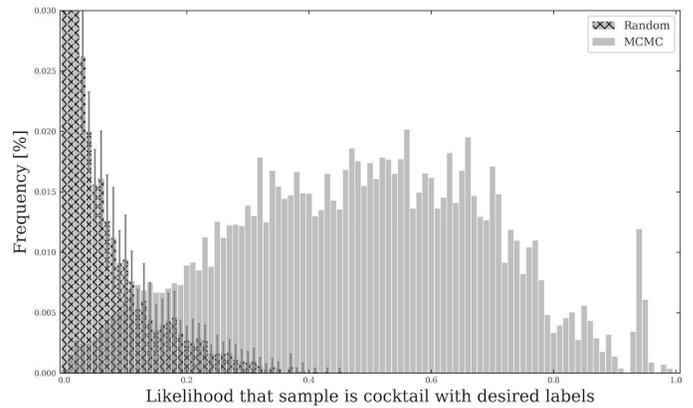}
		\caption{Mean frequencies (\%) of the likelihood of randomly generated samples (hatched bars) and frequencies (\%) of the likelihood of samples generated by MCMC (gray bars). The mean frequencies were calculated from $100\,000$ samples $\times$ 10 iterations. The error bars represent the standard deviations.}
	\label{fig:cocktail_hist}
	\end{figure}

Figure \ref{fig:cocktail_hist} shows the frequencies of $y_{\text{taste}}(\boldsymbol{x})y_{\text{timing}}(\boldsymbol{x})$ occurring for randomly generated samples under the condition of $y_{\text{sparse}}$ and samples generated by the proposed MCMC algorithm. Our MCMC algorithm more efficiently sampled cocktails having large $y_{\text{taste}}(\boldsymbol{x})y_{\text{timing}}(\boldsymbol{x})$ than the random method did. In other words, the MCMC samples were more likely to have the ``Fresh'' and ``All day'' labels. Figure \ref{fig:cocktail_hist} also shows greater variety for the MCMC samples in terms of the measure of the target distribution.

\section{Conclusion}
We have proposed an MCMC algorithm that can generate composition ratio samples satisfying the sparsity and conditions of composition ratios. 
Our empirical results show that the proposed method converges to the target distribution with sparsity conditions, and the combination  with supervised learning can solve a creative problem.
\section*{Acknowledgments}
This research was supported by CREST, JST Grant Number JPMJCR1304, Japan.

\appendix

\Urlmuskip=0mu plus 1mu\relax
\bibliographystyle{named}
\bibliography{arXiv19_obara}

\begin{thebibliography}{}

\bibitem[\protect\citeauthoryear{Andjelkovic}{2018}]{Stevan2019}
Stevan Andjelkovic.
\newblock Cocktails dataset.
\newblock \url{https://github.com/stevana/cocktails}, 2018.

\bibitem[\protect\citeauthoryear{Andrieu \bgroup \em et al.\egroup
  }{2003}]{andrieu2003introduction}
Christophe Andrieu, Nando~D. Freitas, Arnaud Doucet, and Michael~I. Jordan.
\newblock An introduction to {MCMC} for machine learning.
\newblock {\em Machine learning}, 50(1-2):5--43, 2003.

\bibitem[\protect\citeauthoryear{Blei \bgroup \em et al.\egroup
  }{2003}]{blei2003latent}
David~M. Blei, Andrew~Y. Ng, and Michael~I. Jordan.
\newblock Latent {D}irichlet allocation.
\newblock {\em Journal of machine Learning research}, 3(Jan):993--1022, 2003.

\bibitem[\protect\citeauthoryear{Bostrom}{2007}]{bostrom2007estimating}
Henrik Bostrom.
\newblock Estimating class probabilities in random forests.
\newblock In {\em Sixth International Conference on Machine Learning and
  Applications (ICMLA 2007)}, pages 211--216. IEEE, 2007.

\bibitem[\protect\citeauthoryear{Brodie \bgroup \em et al.\egroup
  }{2009}]{brodie2009sparse}
Joshua Brodie, Ingrid Daubechies, Christine D.~Mol, Domenico Giannone, and
  Ignace Loris.
\newblock Sparse and stable {M}arkowitz portfolios.
\newblock {\em Proceedings of the National Academy of Sciences},
  106(30):12267--12272, 2009.

\bibitem[\protect\citeauthoryear{Chib and
  Greenberg}{1995}]{chib1995understanding}
Siddhartha Chib and Edward Greenberg.
\newblock Understanding the metropolis-hastings algorithm.
\newblock {\em The american statistician}, 49(4):327--335, 1995.

\bibitem[\protect\citeauthoryear{Cura}{2009}]{cura2009particle}
Tunchan Cura.
\newblock Particle swarm optimization approach to portfolio optimization.
\newblock {\em Nonlinear analysis: Real world applications}, 10(4):2396--2406,
  2009.

\bibitem[\protect\citeauthoryear{Fern{\'a}ndez and
  G{\'o}mez}{2007}]{fernandez2007portfolio}
Alberto Fern{\'a}ndez and Sergio G{\'o}mez.
\newblock Portfolio selection using neural networks.
\newblock {\em Computers \& Operations Research}, 34(4):1177--1191, 2007.

\bibitem[\protect\citeauthoryear{Geman and Geman}{1984}]{geman1984stochastic}
Stuart Geman and Donald Geman.
\newblock Stochastic relaxation, {G}ibbs distributions, and the {B}ayesian
  restoration of images.
\newblock {\em IEEE Transactions on pattern analysis and machine intelligence},
  6:721--741, 1984.

\bibitem[\protect\citeauthoryear{Hastings}{1970}]{Hastings1970}
Wilfred~K. Hastings.
\newblock Monte {C}arlo sampling methods using {M}arkov chains and their
  applications.
\newblock {\em Biometrika}, 57(1):97--109, 1970.

\bibitem[\protect\citeauthoryear{LeCun \bgroup \em et al.\egroup
  }{2006}]{lecun2006tutorial}
Yann LeCun, Sumit Chopra, Raia Hadsell, MarcAurelio~Aurelio Ranzato, and Fu~J.
  Huang.
\newblock A tutorial on energy-based learning.
\newblock {\em Predicting structured data}, 1(0), 2006.

\bibitem[\protect\citeauthoryear{Markowitz}{1952}]{markowitz1952portfolio}
Harry Markowitz.
\newblock Portfolio selection.
\newblock {\em The journal of finance}, 7(1):77--91, 1952.

\bibitem[\protect\citeauthoryear{Metropolis \bgroup \em et al.\egroup
  }{1953}]{metropolis1953equation}
Nicholas Metropolis, Arianna~W. Rosenbluth, Marshall~N. Rosenbluth, Augusta~H.
  Teller, and Edward Teller.
\newblock Equation of state calculations by fast computing machines.
\newblock {\em The journal of chemical physics}, 21(6):1087--1092, 1953.

\bibitem[\protect\citeauthoryear{Simpson}{1960}]{simpson1960notes}
George~G. Simpson.
\newblock Notes on the measurement of faunal resemblance.
\newblock {\em American Journal of Science}, 258(2):300--311, 1960.

\end{thebibliography}

\end{document}